\documentclass[journal]{IEEEtran}
\usepackage{amsmath,amsfonts}
\usepackage{array}
\usepackage[caption=false,font=normalsize,labelfont=sf,textfont=sf]{subfig}
\usepackage{textcomp}
\usepackage{stfloats}
\usepackage{url}
\usepackage{verbatim}
\usepackage{graphicx}
\usepackage{cite}
\usepackage{amsmath,amssymb,amsfonts}
\usepackage{algorithmic}
\usepackage{xcolor}

\usepackage{multirow}

\usepackage{amsmath, bm} 
\usepackage{amsthm}
\usepackage[ruled,norelsize,linesnumbered]{algorithm2e}
\makeatletter
\newcommand{\removelatexerror}{\let\@latex@error\@gobble}
\makeatother

\DeclareMathOperator*{\Argmin}{argmin}

\newcommand{\tableYes}[0]{{
	\textcolor{green}{\checkmark}
}}
\newcommand{\tableNo}[0]{{
	\textcolor{red}{$\times$}
}}

\newcommand{\bracket}[1]{{
	\left(#1\right)
}}
\newcommand{\squareBracket}[1]{{
	\left[#1\right]
}}
\newcommand{\curlyBracket}[1]{{
	\left\{#1\right\}
}}

\newcommand{\R}{
	\mathbb{R}
}

\newcommand{\N}{
	\mathbb{N}
}

\newcommand{\floor}[1]{
	\lfloor{#1}\rfloor{}
}
\newcommand{\norm}[1]{{
	\left\Vert #1 \right\Vert
}}

\newcommand{\sinc}{
	\operatorname{sinc}
}

\newcommand{\chol}[1]{
	\operatorname{chol} \left( #1 \right)
}

\newcommand{\sphereManifold}[1]{
	\mathbb{S}^{#1}
}

\newcommand{\spdManifold}[1]{
	\mathbb{S}^{#1}_{++}
}

\newcommand{\manifold}[0]{{
		\mathcal{M}
}}
\newcommand{\tangent}[1]{{
		\mathcal{T}_{#1}
}}

\newcommand{\logmap}[2]{{
		\mathbf{Log}_{#2}(#1)
}}
\newcommand{\expmap}[2]{{
		\mathbf{Exp}_{#2}(#1)
}}

\newcommand{\logmapRdSd}[2]{{
		\mathbf{Log}^{\R^{3} \times \sphereManifold{3}}_{#2}\left(#1\right)
}}
\newcommand{\expmapRdSd}[2]{{
		\mathbf{Exp}^{\R^{3} \times \sphereManifold{3}}_{#2}\left(#1\right)
}}
\newcommand{\logmapSD}[2]{{
		\mathbf{Log}^{\sphereManifold{D}}_{#2}\left(#1\right)
}}
\newcommand{\expmapSD}[2]{{
		\mathbf{Exp}^{\sphereManifold{D}}_{#2}\left(#1\right)
}}
\newcommand{\logmapSThree}[2]{{
		\mathbf{Log}^{\sphereManifold{3}}_{#2}\left(#1\right)
}}
\newcommand{\expmapSThree}[2]{{
		\mathbf{Exp}^{\sphereManifold{3}}_{#2}\left(#1\right)
}}
\newcommand{\logmapSPD}[2]{{
		\mathbf{Log}^{\spdManifold{D}}_{#2}\left(#1\right)
}}
\newcommand{\expmapSPD}[2]{{
		\mathbf{Exp}^{\spdManifold{D}}_{#2}\left(#1\right)
}}
\newcommand{\logmapFunction}[1]{{
		\mathbf{Log}_{#1}
}}
\newcommand{\expmapFunction}[1]{{
		\mathbf{Exp}_{#1}
}}

\newcommand{\computationComplexity}[1]{
	\mathcal{O}(#1)
}

\newcommand{\nbSignals}[0]{{
		N
}}
\newcommand{\nbData}[1]{{
		T_{#1}
}}
\newcommand{\nbDataWarped}[1]{{
		Z_{#1}
}}

\newcommand{\dataPosition}[0]{{
		\bm{x}_{\text{pos}}
}}
\newcommand{\dataUnitQuaternion}[0]{{
		\bm{x}_{\text{quat}}
}}

\newcommand{\dataManifold}[1]{{
		\bm{x}_{#1}
}}
\newcommand{\dataManifoldWarped}[1]{{
		\bm{\warped{x}}_{#1}
}}
\newcommand{\dataManifoldMean}[1]{{
		\bm{\warped{y}}_{#1}
}}
\newcommand{\dataManifoldTmp}[1]{{
		\bm{p}_{#1}
}}

\newcommand{\dataTangent}[1]{{
		\bm{u}_{#1}
}}
\newcommand{\dataTangentWarped}[1]{{
		\bm{\warped{u}}_{#1}
}}
\newcommand{\dataTangentMean}[1]{{
		\bm{\warped{v}}_{#1}
}}

\newcommand{\dataOriginalSet}[0]{{
		\mathbf{X}
}}
\newcommand{\dataOriginal}[1]{{
		\mathbf{x}_{#1}
}}

\newcommand{\warped}[1]{{
		\hat{#1}
}}
\newcommand{\dataWarpedSet}[0]{{
		\mathbf{\warped{X}}
}}
\newcommand{\dataWarped}[1]{{
		\mathbf{\warped{x}}_{#1}
}}

\newcommand{\dataMean}[1]{{
		\mathbf{\warped{y}}_{#1}
}}

\newcommand{\warpingFunctionSet}[0]{{
	\mathbf{\Omega}
}}
\newcommand{\warpingFunction}[1]{{
	\mathbf{\omega}_{#1}
}}

\newcommand{\dtwDistance}[2]{{
		\mathcal{D}_{dtw}\left(#1 , #2\right)
}}
\newcommand{\dtwDistanceSymbol}[0]{{
		\mathcal{D}_{dtw}
}}

\newcommand{\distanceMetric}[2]{{
		\mathcal{D}_{#1}(#2)
}}
\newcommand{\distanceComputation}[2]{{
		\text{dist}(#1 , #2)
}}
\newcommand{\optimizationPenaltyTerm}[0]{{
		p(\warpingFunctionSet)
}}
\newcommand{\optimizationPenaltyWeighting}[0]{{
		\lambda
}}
\newcommand{\ttwK}[0]{{
		K
}}
\newcommand{\ntwBasis}[1]{{
		\mathbf{e}_{#1}
}}
\newcommand{\ntwWeights}[1]{{
		\mathbf{\phi}_{#1}
}}
\newcommand{\sincWindowSize}[0]{{
		\nu
}}
\newcommand{\lossWindowSize}[0]{{
	\kappa
}}
\newcommand{\lossStepSize}[0]{{
		\rho
}}
\newcommand{\lossGaussianWeighting}[1]{{
		\hat{g}(#1)
}}
\newcommand{\gaussian}[1]{{
		g(#1)
}}

\newcommand{\dataOriginalTrain}[1]{{
		\mathbf{x}^{\text{train}}_{#1}
}}
\newcommand{\dataOriginalTest}[1]{{
		\mathbf{x}^{\text{test}}_{#1}
}}
\newcommand{\dataWarpedTrain}[1]{{
		\mathbf{\warped{x}}^{\text{train}}_{#1}
}}

\newcommand{\dataMeanTrain}[1]{{
		\mathbf{\warped{y}}^{\text{train}}_{#1}
}}

\makeatletter
	\newcommand\thefontsize{\f@size pt}
\makeatother
\usepackage{printlen}
\uselengthunit{cm}

\usepackage{etoolbox}
\usepackage{capt-of}
\makeatletter
\apptocmd{\@maketitle}{\bigskip\centering\includegraphics[
	trim=0 0 0 0, clip,
	width=\linewidth]{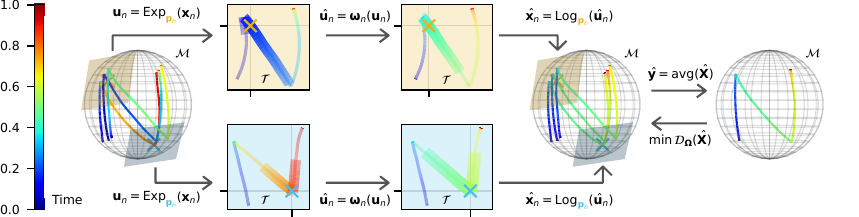}
	\captionof{figure}{
		Schematic overview of RTW: our approach enables the temporal alignment of multiple signals, represented through data on a Riemannian manifold~$\manifold$.
		Given~$\nbSignals$ signals~$\dataOriginal{n}$ with temporal shift (color encoding),
		the exponential and logarithmic maps are utilized to apply time warping within the local tangent spaces~$\tangent{}$.
		RTW utilizes windowed sinc interpolation (bold segments) and parameterized warping functions~$\warpingFunctionSet = \{ \warpingFunction{n} \}_n^\nbSignals$,
		to temporally shift the data within these tangent spaces,
		i.e., $\dataTangent{n}$ becomes $\dataTangentWarped{n}$.
		Finally, the mean signal~$\dataMean{}$ is computed from all warped signals~$\dataWarped{n}$.
		Gradient-based optimization is utilized to find warping functions
		that minimize the distance between all warped signals and their mean.
		See subsection~\ref{sec:elaborationFigureOverviewRTW} for further remarks on this figure.
	}
	\label{fig:overviewRTW}
}{}{}
\makeatother

\begin{document}

\title{
	Riemannian Time Warping:\\
	Multiple Sequence Alignment in Curved Spaces
}

\author{
	Julian Richter$^{1,2}$,
	Christopher A. Erd\H{o}s$^{1}$,
	Christian Scheurer$^{1}$,
	Jochen J. Steil$^{2}$ and
	Niels Dehio$^{1}$%
	\thanks{This work has been submitted to the IEEE for possible publication. Copyright may be transferred without notice, after which this version may no longer be accessible.}
	\thanks{$^{1}$Technology and Innovation Center (TIC), KUKA Deutschland GmbH, Germany, e-mail: {\tt\small Julian.Richter@kuka.com}}%
	\thanks{$^{2}$Institute of Robotics and Process Control (IRP), Technische Universität Braunschweig, Germany}%
}%



\maketitle

\setcounter{footnote}{2}




\begin{abstract}
	Temporal alignment of multiple signals through time warping is crucial in many fields,
	such as classification within speech recognition or robot motion learning.
	Almost all related works are limited to data in Euclidean space. 
	Although an attempt was made in 2011 to adapt this concept to unit quaternions, 
	a general extension to Riemannian manifolds remains absent.
	Given its importance for numerous applications in robotics and beyond, 
	we introduce Riemannian Time Warping~(RTW).
	This novel approach efficiently aligns multiple signals 
	by considering the geometric structure of the Riemannian manifold in which the data is embedded.
	Extensive experiments on synthetic and real-world data, 
	including tests with an LBR iiwa robot, 
	demonstrate that RTW consistently outperforms state-of-the-art baselines in both averaging and classification tasks.
\end{abstract}




\section{Introduction}


Time warping describes the temporal alignment of several signals 
-- a fundamental operation for processing sequential data commonly used
in speech recognition~\cite{
	Shimodaira_DynamicTimeAlignmentKernelInSupportVectorMachine_2002,
	Muda_VoiceRecognitionAlgorithmsUsingMFCCandDTWTechniques_2010
}
and
human action recognition~\cite{
	Sheikh_ExploringTheSpaceOfaHumanAction_2005,
	Sempena_HumanActionRecognitionUsingDynamicTimeWarping_2011
}.
In robotics, temporal alignment is crucial for learning movement primitives:
it helps significantly to extract a consistent motion pattern from multiple human demonstrations
that are naturally provided with temporal variations.
Given multiple signals with varying lengths describing the same underlying process,
each signal is warped with the goal to improve their temporal alignment~\cite{
	Steil_TaskLevelImitationLearningUsingVarianceBasedMovementOptimization_2009,
	Calinon_GMM_2007
}.

Dynamic Time Warping (DTW)~\cite{Sakoe_DTW_1978}
finds optimal alignment through Dynamic Programming~\cite{Bellman_DynamicProgramming_1954}
but suffers from exponential complexity that prevents application to larger datasets.
Alternative algorithms sacrifice global optimality 
to accelerate the search process by several orders of magnitude~\cite{
	Zhou_GTW_2012,
	Khorram_TTW_2019,
	Kawano_NTW_2020
}.

Almost all related works focus on Euclidean data.
However, many robotics problems
involve temporal phenomena on Riemannian manifolds (cf. Figure~\ref{fig:overviewRTW}).
For example, demonstrated trajectories naturally lie on the unit sphere $\sphereManifold{3}$ for orientations
or on symmetric positive definite matrices $\spdManifold{D}$ describing a $D$-dimensional stiffness or inertia.
Applying Euclidean time warping can lead to geometrically invalid interpolations 
(e.g., averaging quaternions in Euclidean space yields non-unit vectors)
and suboptimal alignment that ignores geometric constraints~\cite{Calinon_GaussiansOnRiemannianManifoldsForRobotLearningAndAdaptiveControl_2020}.
While Quaternion Dynamic Time Warping (QDTW)~\cite{Jablonski_QDTW_2012}
extends DTW to unit quaternions, it retains exponential complexity.
To the best of our knowledge, no generic time warping method handles structured non-Euclidean data,
partly because sum and scalar operations used in related works 
are not defined on Riemannian manifolds~\cite{Lee_RiemannianManifoldsAnIntroductionToCurvature_2006}.

As our main contribution,
we present \emph{Riemannian Time Warping~(RTW)} 
for efficient temporal alignment of multiple signals with varying lengths
while explicitly considering the Riemannian manifold~$\manifold$ on which the data resides.
We evaluate RTW on datasets in $\R^D$, $\sphereManifold{D}$ and $\spdManifold{D}$.
RTW consistently outperforms state-of-the-art methods~\cite{Khorram_TTW_2019, Kawano_NTW_2020} 
w.r.t. alignment quality and mean representation at linear complexity,
and provides significant benefits for learning movement primitives.
We validate our approach on real-robot experiments using a KUKA LBR iiwa manipulator
in~$\R^{3} \times \sphereManifold{3}$.
A video is available at {\tt\small https://youtu.be/v\_k2LwEDH9I}. 

Notation:
In the remainder,
we will use square bracket notation for discrete values,
e.g., $\dataOriginal{}[t] \in \R^D, \dataManifold{}[t] \in \manifold$,
and round bracket notation for continuous functions,
e.g., $\ntwWeights{}(z)$.


\section{Related Work}


Most related works focus on aligning $\nbSignals = 2$ signals of length~$\nbData{}$.
Despite its age, DTW~\cite{Sakoe_DTW_1978} remains widely used due to its simplicity and effectiveness.
Extending DTW to $\nbSignals > 2$ signals introduces significant computational challenges.
Multiple Multi-Dimensional Dynamic Time Warping (MMDDTW)~\cite{Parinya_MMDDTW_2012}
computes the distance matrix over all $\nbSignals$ dimensions.
While globally optimal through Dynamic Programming,
it becomes unfeasible for $N \gg 2$ due to complexity~$\computationComplexity{\nbData{}^{\nbSignals}}$.
The Non-Linear Alignment and Averaging Framework (NLAAF)~\cite{GUPA_NLAAF_1996}
iteratively aligns pairs of signals while adjusting previous solutions,
with complexity~$\computationComplexity{\nbSignals \nbData{}^{2}}$.

Generalized Time Warping (GTW)~\cite{Zhou_GTW_2012},
Trainable Time Warping (TTW)~\cite{Khorram_TTW_2019},
and Neural Time Warping (NTW)~\cite{Kawano_NTW_2020}
are newer methods that align several signals in linear time.
GTW uses pre-defined basis functions, which is cumbersome for complex warpings.
TTW and NTW transform discrete signals into continuous time via sinc interpolation, 
apply an index shift, and sample back at $Z$ uniformly distributed time points.
TTW estimates discrete sine transform coefficients, while NTW uses a neural network.
Both outperform GTW~\cite{Khorram_TTW_2019, Kawano_NTW_2020}.

To the best of our knowledge,
QDTW~\cite{Jablonski_QDTW_2012}
is the only warping scheme that explicitly deals with a Riemannian manifold,
focusing exclusively on unit quaternions.
However, in recent years, Riemannian manifolds have gained popularity in various domains --
especially in robotics~\cite{
	Jaquier_GeometryAwareManipulabilityLearningTrackingAndTransfer_2021,
	Jaquier_RiemannianGeometryAsAUnifyingTheoryForRobotMotionLearningAndControl_2022,
	Saveriano_LearningStableRoboticSkillsOnRiemannianManifolds_2023,
	Loew_GAFROGeometryAlgebraForRobotics_2024
},
as they provide a convenient way to generalize methods originally developed for the Euclidean space.
This letter presents the first approach for time warping on generic Riemannian manifolds,
that applies to datasets with $\nbSignals \gg 2$.
Table~\ref{tab:overviewComputationalComplexity} compares related works.

\begin{table}[!b]
	\centering
	\caption{Comparison between state-of-the-art Time Warping Methods}
	\label{tab:overviewComputationalComplexity}
	\renewcommand{\arraystretch}{1.2}
	\begin{tabular}{ c | c c c c }
		Method
		& $\nbSignals $ & $\sphereManifold{3}$ & $\manifold$
		& Complexity \\
		\hline
		DTW~\cite{Sakoe_DTW_1978}
		& $=\,2$           & \tableNo \,             & \tableNo
		& $ \computationComplexity{ \nbData{}^2 }$ \\
		QTDW~\cite{Jablonski_QDTW_2012}
		& $=\,2$           & \tableYes \,            & \tableNo 
		& $ \computationComplexity{ \nbData{}^2 } $ \\
		MMDDTW~\cite{Parinya_MMDDTW_2012}
		& $>\,2$       & \tableNo \,             & \tableNo
		& $ \computationComplexity{ \nbData{}^\nbSignals } $ \\
		TTW~\cite{Khorram_TTW_2019}
		& $\gg 2$       & \tableNo \,             & \tableNo
		& $ \computationComplexity{ \nbSignals \nbDataWarped{} } $ \\
		NTW~\cite{Kawano_NTW_2020}
		& $\gg 2$       & \tableNo \,             & \tableNo
		& $ \computationComplexity{ \nbSignals \nbDataWarped{} \nbData{} } $ \\
		RTW~[proposed]
		& $\gg 2$       & \tableYes \,            & \tableYes
		& $ \computationComplexity{ \nbSignals \nbDataWarped{} } $ \\
	\end{tabular}
\end{table}


\section{Time Warping in the Continuous Time Domain}


Here we first introduce a simplified version of our approach dedicated to Euclidean data $\dataOriginal{} \in \R^D$.
The full \emph{Riemannian Time Warping (RTW)} algorithm considering curved spaces with $\dataManifold{} \in \manifold$ is then proposed in the next section.


\subsection{Problem Statement}


Consider $\nbSignals$ discrete signals
$
	\dataOriginalSet
	=
	\{
	\dataOriginal{n}
	\}_{n=1}^{N}
$
of lengths $\nbData{n}$,
i.e. 
$
	\dataOriginal{n}
	= 
	\{
	\dataOriginal{n}[t]
	\}_{t=1}^{\nbData{n}}
$,
where $\dataOriginal{n}[t] \in \R^{D}$.
Signals with temporal modifications
$
	\dataWarpedSet
	=
	\{
	\dataWarped{n}
	\}_{n=1}^{N}
$
of new length $\nbDataWarped{} \geq \nbData{\text{max}} = \max \curlyBracket{ \nbData{n} }_{n=1}^{\nbSignals}$,
with
$
	\dataWarped{n} =     
	\{
	\dataWarped{n}[z]
	\}_{z=1}^{Z}
$
and
$\dataWarped{n}[z] \in \R^{D}$
are obtained by applying a set of parameterized warping functions
$
\warpingFunctionSet = \curlyBracket{
	\warpingFunction{n} :
	\curlyBracket{1, \cdots, Z}
	\mapsto 
	[0,1]
}_{n=1}^{N}
$.	
The \emph{Multiple Sequence Alignment} problem consists of
\begin{equation}
	\label{eq:msaProblem}
	\warpingFunctionSet_{opt}
	=
	\Argmin_{\warpingFunctionSet}
	\,
	\distanceMetric{\warpingFunctionSet}{ \dataWarpedSet }
	\quad
	\text{s.t.}
	\quad
	\eqref{eq:constraintBoundary},
	\eqref{eq:constraintContinuous},
	\eqref{eq:constraintMonton}
	\, ,
\end{equation}
where the warping functions $\warpingFunction{n}$ must satisfy three constraints
\begin{subequations}
	\begin{alignat}{2}
		\label{eq:constraintBoundary}
		&\text{Boundary Constraint:}
		&\quad
		&\warpingFunction{n}[1] = 0, 
		\warpingFunction{n}[Z] = 1
		\\
		\label{eq:constraintContinuous}
		&\text{Continuity Constraint:}
		&\quad
		&\warpingFunction{n}[z+1] - \warpingFunction{n}[z]
		\leq
		\frac{1}{\nbData{\text{max}}}
		\\
		\label{eq:constraintMonton}
		&\text{Monotonicity Constraint:}
		&\quad
		&\warpingFunction{n}[z]
		\leq
		\warpingFunction{n}[z+1]
	\end{alignat}
\end{subequations}
and~$\distanceMetric{\warpingFunctionSet}{ \dataWarpedSet }$ is a distance measure
between all warped signals~$\dataWarped{n}$ and their mean
\begin{equation}
	\label{eq:meanComputation}
	\dataMean{}
	=
	\frac{1}{N} \textstyle \sum_{n=1}^{N} \dataWarped{n}
	\, .
\end{equation}
The challenge is to find optimal and feasible warping functions efficiently.
This is non-trivial, as $\dataOriginalSet$ is a set of discrete signals.

	
\subsection{Proposed Time Warping Algorithm in $\R^{D}$}
\label{sec:rtwMethodEuclidean}
	

We transform~$\dataOriginalSet$
into the continuous time domain
to compute the aligned signals~$\dataWarpedSet$
through sinc interpolation\footnote{%
	Sinc interpolation provides optimal reconstruction for signals sampled at or above Nyquist rate, i.e., twice the maximum frequency of the data~\cite{Shannon_CommunicationInThePresenceOfNoise_1949}.
}
\begin{equation}
	\label{eq:sincInterpolationWindowed}
	\dataWarped{n}[z]
	=
	\sum
	_{m = \floor{\warpingFunction{n}[z] \nbData{n}} - \sincWindowSize}
	^{    \floor{\warpingFunction{n}[z] \nbData{n}} + \sincWindowSize}
	\dataOriginal{n}[m]
	\sinc(m - \warpingFunction{n}[z] \,\, \nbData{n})
	\, ,
\end{equation}
where $\floor{\cdot}$ denotes the floor-function
and $\sincWindowSize \in \N_{0}$ is the window size.
In contrast to NTW~\cite{Kawano_NTW_2020},
we adapt a windowed sinc interpolation\footnote{%
	We apply clamping to always receive valid indices for the full interpolation window.
	Given a data point $\dataOriginal{n}[m]$,
	we set $m = \max(1, \min(\nbData{n}, m))$.
}
as in TTW~\cite{Khorram_TTW_2019},
resulting in a computational complexity of
$
\computationComplexity{
	\nbSignals \nbDataWarped{}
}
$.
Since the amplitudes in the $\sinc$ function decay over time,
$\sincWindowSize = 10$
captures more than 99\% of its power~\cite{Khorram_TTW_2019}.
Similar to NTW~\cite{Kawano_NTW_2020},
we utilize $\nbSignals$ orthogonal basis vectors $\ntwBasis{k} \in \R^{\nbSignals}$
that are derived from the Gram-Schmidt process or QR-decomposition with
$\ntwBasis{1} = \frac{1}{\sqrt{\nbSignals}} \left[1, \cdots, 1\right]^T$,
and model the warping functions~$\warpingFunction{n}$ through
\begin{equation}
	\label{eq:warpingFunctionNTW}
	\warpingFunction{}_{n}[z]
	=
	\hat{z} \sqrt{\nbSignals} \ntwBasis{1}
	+
	\hat{z} \bracket{1 - \hat{z}}
	\sum_{k=1}^{N-1}
	\squareBracket{
		\ntwWeights{\theta}\left(\hat{z}\right)
	}_{k}
	\,
	\ntwBasis{k+1}
	\, ,
\end{equation}
where the $\theta$-parameterized function
$
\ntwWeights{\theta}
:
\squareBracket{0,1}
\mapsto
\R^{\nbSignals-1}
$
is modeled as a neural network,
and $\hat{z} = \frac{z-1}{Z-1}$.
The boundary constraint~\eqref{eq:constraintBoundary} is guaranteed by design.
During warping,
indices of a signal may be duplicated to improve the temporal alignment,
requiring a longer signal,
i.e.,
$Z \geq \nbData{\text{max}} \,$.
Hence,
the continuity constraint~\eqref{eq:constraintContinuous}
is fulfilled with $Z=\nbSignals\nbData{\text{max}}$~\cite{Kawano_NTW_2020}.
To enforce the monotonicity constraint~\eqref{eq:constraintMonton},
a $\optimizationPenaltyWeighting$-weighted penalty term is added to the loss as in~\cite{Kawano_NTW_2020}
\begin{equation}
	\label{eq:msaProblemRTW}
	\textstyle
	\warpingFunctionSet_{opt}
	=
	\Argmin_{\warpingFunctionSet}
	\bracket{
		\distanceMetric{\warpingFunctionSet}{ \dataWarpedSet }
		+
		\optimizationPenaltyWeighting \, \optimizationPenaltyTerm
	}
	\, ,
\end{equation}
\begin{equation}
	\optimizationPenaltyTerm
	=
	\sum_{n=1}^{\nbSignals}
	\sum_{z=1}^{Z-1}
	\max\bracket{
		\warpingFunction{n}[z] - \warpingFunction{n}[z+1]
		\, ,
		0
	}
	\, .
\end{equation}
The unconstrained optimization problem~\eqref{eq:msaProblemRTW} approximates~\eqref{eq:msaProblem} 
and is solved using state-of-the-art optimization techniques.

\begin{figure}[!t]
	\centering
	\includegraphics[
	trim=0 5 0 5, clip,
	width=\linewidth]{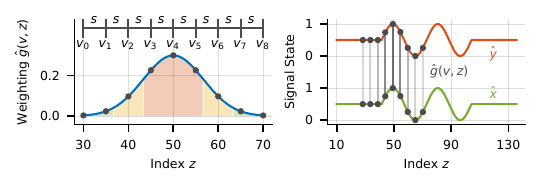}
	\caption{
		Visualization of the Gaussian-weighted window loss
		between two signals~$\dataWarped{}$ (green) and~$\dataMean{}$ (red)
		as defined in~\eqref{eq:gaussianWeightedWindowLoss}.
		The example shows one segment at
		index~$z = 50$ with
		window size~$\lossWindowSize = 4$
		and step size~$\lossStepSize = 5$,
		where red, yellow and green mark the areas covered by
		one, two and three times the variance~$\sigma$ of the Gaussian.
		By considering a segment around index~$z$,
		the first peak of both signals becomes distinguishable from the second.
	}
	\label{fig:gaussianWeightedLoss}
\end{figure}

Finally,
we define the distance measure~$\distanceMetric{\warpingFunctionSet}{ \dataWarpedSet }$
as a Gaussian-weighted window loss of point segments
\begin{equation}
	\begin{gathered}
		\label{eq:gaussianWeightedWindowLoss}
		\distanceMetric{\warpingFunctionSet}{ \dataWarpedSet }
		=
		\frac{1}{\nbSignals \nbDataWarped{}}
		\sum_{n=1}^{\nbSignals}
		\sum_{z=1}^{\nbDataWarped{}}
		\sum_{k=-\lossWindowSize}^{\lossWindowSize}
		\lossGaussianWeighting{v,z} \,
		\distanceComputation{\dataWarped{n}[v]}{\dataMean{}[v]}
		\, ,
		\\
		\lossGaussianWeighting{v,z}
		=
		\frac{
			\gaussian{v}
		}{
			\sum_{i=-\lossWindowSize}^{\lossWindowSize}
			\gaussian{z + i \lossStepSize}
		}
		\,\,\,\,\, , \,\,\,\,\,
		\gaussian{j}
		=
		\exp \bracket{
			-\frac{j^2}{2 \sigma^2}
		}
		\, ,
	\end{gathered}
\end{equation}
with $v = z + k \lossStepSize$,
where
the window size~$\lossWindowSize \in \N_{0}$
describes how many additional indices are considered at index $z$
and
the step size~$\lossStepSize \in \N$
defines the distance between them,
resulting in the indices~$v$
that form a segment\footnote{\label{fn:clampingSegmentIndices}%
	As for the sinc interpolation,
	we apply clamping to always receive valid indices for a full segment,
	i.e., we set $v = \max(1, \min(\nbDataWarped{}, v))$.
}.
Each index within a segment
is weighted with the Gaussian kernel $\lossGaussianWeighting{v,z}$.
Given the empirical rule,
we set $\sigma = \frac{\lossWindowSize \lossStepSize}{3} + \epsilon$
with a small $\epsilon > 0$ for numerical reasons
to ensure a weighting,
such that all indices lie within $99,7\%$ of the Gaussian's power,
emphasizing central indices
as visualized in Figure~\ref{fig:gaussianWeightedLoss}.
Note that by setting $\lossWindowSize = 0$,
we receive a point-to-point metric
similar to TTW or NTW.
For data in $\R^{D}$,
we utilize the Euclidean distance metric
\begin{equation}
	\label{eq:distanceMetricEuclidean}
	\text{dist}_{\R^D}(\dataWarped{}[z] , \dataMean{}[z])
	=
	\norm{ \dataWarped{}[z] - \dataMean{}[z] }
	\, .
\end{equation}

\section{RTW Extension for Riemannian Manifolds}


Extending RTW to data on a Riemannian manifold $\manifold$
is not straightforward as
there is no sum and scalar multiplication defined,
i.e., it is not a vector space.
Accordingly, the classical time warping approaches yield geometrically invalid results,
because the warped signals and their mean would not reside on the manifold.
Specifically, 
sinc interpolation~\eqref{eq:sincInterpolationWindowed},
mean computation~\eqref{eq:meanComputation}
and distance evaluation~\eqref{eq:distanceMetricEuclidean}
would provide outcomes that do not respect the manifold structure. 
To address this, our key idea is to utilize Euclidean tangent spaces~$\tangent{}$
that locally preserve distances. 
Then, we adjust the
sinc interpolation (see~\ref{sec:RiemannianExtensionInterpolation}),
mean computation (see~\ref{sec:RiemannianExtensionMean})
and distance evaluation (see~\ref{sec:RiemannianExtensionDistanceMeasure}),	
to incorporate the curved manifold structure by operating within the tangent spaces. 
These novel modifications enable temporal alignment in the continuous time domain to be applied on Riemannian manifolds
without affecting the computational complexity
$
\computationComplexity{
	\nbSignals \nbDataWarped{}
}
$:
RTW scales linearly with $\nbSignals$ and $\nbDataWarped{}$.

More specifically,
a manifold~$\manifold$ is a $D$-dimensional smooth space~\cite{Lee_RiemannianManifoldsAnIntroductionToCurvature_2006},
for which there exists a tangent space~$\tangent{\dataManifoldTmp{}}$
for each point~$\dataManifoldTmp{} \in \manifold$ on the manifold,
with $\dataManifoldTmp{}$ being the origin of~$\tangent{\dataManifoldTmp{}}$.
A manifold with a positive definite inner product
defined on the tangent space is called Riemannian.
The exponential map~$\expmapFunction{\dataManifoldTmp{}}: \tangent{\dataManifoldTmp{}} \mapsto \manifold$
maps a point
from the tangent space onto the manifold.
The inverse operation
is the logarithmic map~$\logmapFunction{\dataManifoldTmp{}} : \manifold \mapsto \tangent{\dataManifoldTmp{}}$.
These mappings are distance-preserving,
i.e., the geodesic distance between the point~$\dataManifoldTmp{}$
and any other point is maintained.
This feature allows indirect computation to be performed on the manifold
by applying Euclidean operations in the tangent space. 
The formulation of the exponential and logarithmic map is manifold dependent.


\subsection{Windowed Sinc Interpolation on Riemannian Manifolds}
\label{sec:RiemannianExtensionInterpolation}


The windowed sinc interpolation in~\eqref{eq:sincInterpolationWindowed} for a time index~$z$
constitutes a weighted sum,
which is not a closed operation under $\manifold$.
Therefore,
we project the original $2 \sincWindowSize + 1$ data points from the manifold 
onto the tangent space~$\tangent{\dataManifoldTmp{n}[z]}$
through 
$
	\dataTangent{n,z}[m]
	=
	\logmap{\dataManifold{n}[m]}{\dataManifoldTmp{n}[z]}
$,
where we warp and interpolate
\begin{equation}
	\label{eq:sincInterpolationManifoldTangent}
	\dataTangentWarped{n}[z]
	=
	\sum
	_{m = \floor{\warpingFunction{n}[z] \nbData{n}} - \sincWindowSize}
	^{    \floor{\warpingFunction{n}[z] \nbData{n}} + \sincWindowSize}
	\dataTangent{n,z}[m]
	\sinc(m - \warpingFunction{n}[z] \, \nbData{n})
	\, .
\end{equation}
Then, the result for the time index~$z$ is projected
from the specific tangent space $\tangent{\dataManifoldTmp{n}[z]}$
back onto the manifold via
\begin{equation}
	\label{eq:sincInterpolationManifoldWarped}
	\dataManifoldWarped{n}[z]
	=
	\expmap{
		\dataTangentWarped{n}[z]
	}{\dataManifoldTmp{n}[z]}
	\, .
\end{equation}
Note that the choice of the tangent point~$\dataManifoldTmp{n}[z] \in \manifold$
is important to minimize undesired distortions~\cite{Jaquier_UnravelingTheSingleTangentSpaceFallacyAnAnalysisAndClarificationForApplyingRiemannianGeometryInRobotLearning_2024}.
Therefore, we initialize the origin of the tangent space at the center of the interpolation window as
$
	\dataManifoldTmp{n}[z]
	=
	\dataManifold{n}\squareBracket{ \, \floor{ \warpingFunction{n}[z] \, \nbData{n} } \, }
$,
and refine it through few Gauss-Newton iterations
$
	\dataManifoldTmp{n}[z]
	\leftarrow
	\dataManifoldWarped{n}[z]
$
to approximate the optimal tangent point that minimizes distortions,
i.e.,
we compute a weighted Fréchet mean within the interpolation window.
Due to the continuity constraint of the warping functions~$\warpingFunction{n}$,
the Fréchet mean is constrained to lie on the sinc interpolated line
connecting the three adjacent discrete data points
$\dataManifold{n}[\floor{\warpingFunction{n}[z] \nbData{n}}-1]$, $\dataManifold{n}[\floor{\warpingFunction{n}[z] \nbData{n}}]$ and $\dataManifold{n}[\floor{\warpingFunction{n}[z] \nbData{n}}+1]$.
Further note that the optimal tangent space
is reevaluated for each time index~$z$ and signal~$n$,
instead of projecting all data points onto one single tangent space.


\subsection{Signal Mean Computation on Riemannian Manifolds}
\label{sec:RiemannianExtensionMean}


Similar to the sinc interpolation,
the mean computation~\eqref{eq:meanComputation} is adjusted
by utilizing Gauss-Newton iterations.
Starting from an initial estimate $\dataManifoldMean{}[z] \in \manifold$,
e.g.,
$\dataManifoldMean{}[z] = \dataManifoldWarped{1}[z]$,
the $z$-th data point~$\curlyBracket{ \dataManifoldWarped{n}[z] \in \manifold }_{n=1}^{\nbSignals}$
of each signal
is projected onto the tangent space~$\tangent{\dataManifoldMean{n}}$
through $\dataTangentWarped{n}[z] = \logmap{\dataManifoldWarped{n}[z]}{\dataManifoldMean{}[z]}$.
We obtain the center of these projections as
\begin{equation}
	\label{eq:GaussNewtonTangentMean}
	\dataTangentMean{}[z]
	= 
	\frac{1}{\nbSignals} 
	\sum_{n=1}^{\nbSignals} 
	\dataTangentWarped{n}[z]
	\, .
\end{equation}
This center is projected back
from the tangent space $\tangent{\dataManifoldMean{}[z]}$
onto the manifold $\manifold$
to update the mean at time index $z$ through
\begin{equation}
	\label{eq:GaussNewtonManifoldMeanUpdate}
	\dataManifoldMean{}[z]
	\leftarrow
	\expmap{\dataTangentMean{}[z]}{\dataManifoldMean{}[z]}
	\, .
\end{equation}
This process,
\eqref{eq:GaussNewtonTangentMean} and~\eqref{eq:GaussNewtonManifoldMeanUpdate},
is repeated until convergence,
which is typically reached after few iterations,
e.g., see~\cite{Calinon_GaussiansOnRiemannianManifoldsForRobotLearningAndAdaptiveControl_2020}.


\subsection{Distance Computation on Riemannian Manifolds}
\label{sec:RiemannianExtensionDistanceMeasure}


The distance metric~\eqref{eq:distanceMetricEuclidean} is adapted as well.
We utilize the geodesic distance on Riemannian manifolds
\begin{equation}
	\label{eq:distanceMetricRiemannian}
	\text{dist}_{\manifold}(\dataManifoldWarped{}[z] , \dataManifoldMean{}[z])
	=
	\norm{ \logmap{\dataManifoldWarped{}[z]}{\dataManifoldMean{}[z]} }_{\dataManifoldMean{}[z]}
	\, ,
\end{equation}
where the norm is computed with respect to the Riemannian metric at $\dataManifoldMean{}[z]$,
which depends on the manifold's geometry.


\subsection{Further Remarks on the Overview provided in Fig.~\ref{fig:overviewRTW}}
\label{sec:elaborationFigureOverviewRTW}


Figure~\ref{fig:overviewRTW} provides a schematic overview, summarizing the proposed RTW approach.
Three similar signals with significant temporal shifts are considered in~$\sphereManifold{2}$,
a unit sphere in~$\R^3$.
The color gradient from blue to red indicates the progression of time along the trajectories. 
We illustrate the mapping of one signal onto the tangent space for two different origins~$\dataManifoldTmp{n}[z]$, 
where the tangent space is a two-dimensional plane. 
As expected, distortion is noticeable only for data points far from the origin.
For visualization purposes, 
we choose a large interpolation window~$\sincWindowSize$, indicated by the bold segments.
Note that these parts are close to the origin of the tangent space but not spatially centered around it, 
as the window is centered in the time domain instead.
Furthermore,
RTW only projects data points from a single signal onto each tangent space during interpolation,
while related works often project all data points onto them,
e.g., see~\cite{
	Calinon_AnApproachForImitationLearningOnRiemannianManifolds_2017,
	Calinon_GaussiansOnRiemannianManifoldsForRobotLearningAndAdaptiveControl_2020,
	Saveriano_LearningStableRoboticSkillsOnRiemannianManifolds_2023
}.
Applying warping functions~$\warpingFunction{n}$ does not alter the shape of a signal in the tangent spaces; 
rather, signals are temporally adjusted, as shown by the new color coding. 
RTW results in three signals on the manifold that are well aligned. 
After a few Gauss-Newton iterations, we obtain the signal mean, respecting the unit sphere structure.


\section{Simulations and Experiments}


This section compares the proposed RTW approach with baselines
in various settings considering datasets in 
$\R^{D}$, $\sphereManifold{D}$ and $\spdManifold{D}$.
We also apply RTW as a pre-processing step for learning movement primitives
for a real-robot teaching scenario in~$\R^{3} \times \sphereManifold{3}$
using an LBR~iiwa manipulator.

RTW, TTW and NTW have been implemented in PyTorch,
utilizing the Adam optimizer
with a learning rate of $0.01$
and Autograd
for gradient computations.
The best model over 256 epochs was selected during the comparisons.
For RTW,
we use a neural network to model $\ntwWeights{\theta}(\hat{z})$
with four fully connected linear layers\footnote{
	We also evaluated simpler model architectures,
	such as estimating coefficients of a polynomial,
	which achieved comparable results at faster runtime for our experiments.
	However, for generalization purposes, and to maintain a fair comparison with NTW,
	we stick to the neural network architecture.
},
ReLU activations
and skip connections
of size 1-512-512-1025-(N-1),
as suggested in~\cite{Kawano_NTW_2020}.
However, unlike NTW, 
we do not initialize the weights with zero, 
as preliminary trials have shown that this significantly degrades the performance.
Instead, we apply symmetry breaking~\cite{Tanaka_NoethersLearningDynamicsRoleOfSymmetryBreakingInNeuralNetworks_2021},
initializing weights with the Xavier initialization method~\cite{Xavier_UnderstandingTheDifficultyOfTrainingDeepFeedforwardNeuralNetworks_2010}.

We use a window size of $\sincWindowSize = 10$
for the sinc interpolation in RTW/TTW~\eqref{eq:sincInterpolationManifoldTangent}.
For the Gaussian-weighted window loss in RTW~\eqref{eq:gaussianWeightedWindowLoss},
we choose
window size~$\lossWindowSize = 5$
and
step size~$\lossStepSize = 5$,
which had negligible impacts on the runtime.
The penalty term~\eqref{eq:msaProblemRTW} for RTW/NTW is weighted with~$\optimizationPenaltyWeighting = 100$.

The number of learnable parameters~$\theta$ in NTW and RTW
is $263680 + 1025 (\nbSignals-1) + (\nbSignals-1)$,
i.e., the number of weights and biases from all layers of the neural network,
whereas it is only $\nbSignals \ttwK$ for TTW due to the design of the warping function
\begin{equation}
	\label{eq:warpingFunctionTTW}
	\warpingFunction{}_{n}^{ttw}[z]
	=
	\hat{z}
	+
	\sum_{k=1}^{\ttwK}
	\alpha_{n,k} \sin \left( \pi k \hat{z} \right)
	\quad , \quad
	\hat{z} = \frac{z-1}{Z-1}
	\, ,
\end{equation}
where $\ttwK \in \N_{+}$ defines how many components of the discrete sine tranform are used
to model the warping functions.

We also recommend watching the accompanying video.


\subsection{Extensive Benchmark on the UCR Time Series Archive}


We evaluate the performance of the simplified RTW (as presented in~\ref{sec:rtwMethodEuclidean})
w.r.t. TTW/NTW in~$\R^1$.
Therefore, we perform \textit{DTW averaging} and \textit{classification tasks}
on the UCR Time Series Archive~\cite{Dau_UCRArchive_2018},
replicating the experiments in~\cite{Khorram_TTW_2019, Kawano_NTW_2020}
to enable consistent benchmarking and fair comparisons.
This archive consists of $128$~datasets covering various applications;
each is pre-split into train~$\dataOriginalTrain{n,c}$
and test data~$\dataOriginalTest{n,c}$,
containing $\nbSignals \in \curlyBracket{16,\ldots,16800}$ signals
classified into $c \in \curlyBracket{2,\ldots,60}$ classes,
with $\nbData{} \in \curlyBracket{15,\ldots,2844}$ data points,
where most signals are of the same length,
i.e., $\nbData{1} = \ldots = \nbData{\nbSignals}$.


\subsubsection{DTW Averaging Task}


For all datasets,
we perform temporal alignment on the training data~$\dataOriginalTrain{n,c}$
for each class~$c$ separately,
to estimate its mean signal~$\dataMeanTrain{c}$.
We then compute the accumulated DTW distance
$
\sum_{n=1}^{N} \dtwDistance{\dataWarpedTrain{n,c}}{\dataMeanTrain{c}}
$
\cite{Sakoe_DTW_1978}
for all three methods and perform a paired t-test 
($\alpha = 0.05$),
to find statistically significant differences.
Table~\ref{tab:ucrResults}-1. lists the results,
where each entry describes
for how many datasets a method significantly outperformed the other:
RTW and NTW are comparable,
whereas RTW clearly outperforms TTW.

\begin{table}[!b]
	\centering
	\caption{Statistical Comparison on the UCR Time Series Archive}
	\label{tab:ucrResults}
	\renewcommand{\arraystretch}{1.2}
	\begin{tabular}{c | c | c }
		& \!\!\!\!  1. Averaging    & 2. Classification \!\!\!\! \\
		\hline
		\!\!\!\!  TTW significantly better than RTW &          7.34\%  &          5.47\%  \\
		\!\!\!\!  RTW significantly better than TTW & \textbf{39.84\%} & \textbf{54.69\%} \\
		\!\!\!\!  No significant difference         &         52.82\%  &         39.84\%  \\
		\hline
		\!\!\!\!  NTW significantly better than RTW &          7.81\%  &          6.25\%  \\
		\!\!\!\!  RTW significantly better than NTW & \textbf{ 8.59\%} & \textbf{35.16\%} \\
		\!\!\!\!  No significant difference         &         83.60\%  &         58.59\%  \\
	\end{tabular}
\end{table}


\subsubsection{Classification Task}

\begin{figure}[!t]
	\centering
	\includegraphics[
		trim=0 0 0 8, clip,
		width=\linewidth]{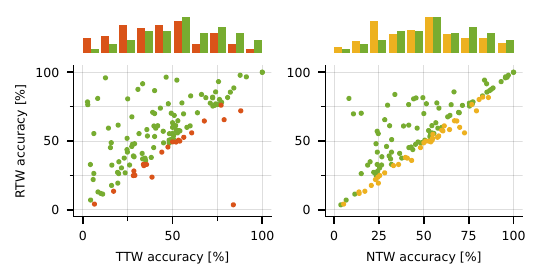}
	\caption{
		Classification performance of TTW (red), NTW (yellow), RTW (green)
		on all 128 datasets
		from the UCR Time Series Archive.
		Each point in the chart represents the TTW/NTW accuracy (x-axis) and RTW accuracy (y-axis).
		Due to overlapping points,
		a histogram shows the accuracy-distribution at the top.
	}
	\label{fig:ucrClassificationResults}
\end{figure}

We conduct classification tasks
on the remaining test data~$\dataOriginalTest{n,c}$.
Given the estimated means~$\dataMeanTrain{c}$
for each class $c$
of a dataset,
we predict the class label~$\hat{c}_{n}$
for all test signals~$\dataOriginalTest{n,c}$
utilizing nearest centroid classification as
$
\hat{c}_{n}
=
\Argmin_{c}
\mathcal{D}_{dtw}(\dataOriginalTest{n,c} , \dataMeanTrain{c})
$.
The results of the paired t-tests ($\alpha = 0.05$)
between TTW, NTW and RTW
are shown in Table~\ref{tab:ucrResults}-2.
RTW learns complex non-linear warpings,
outperforming both TTW and NTW significantly.
Compared to a neural network,
modeling the warping functions
through the discrete sine transform
as in TTW~\eqref{eq:warpingFunctionTTW} limits their flexibility.
Without applying symmetry breaking in NTW, 
the neural network only learns a linear model, 
which limits its warping functions to a quadratic structure.
Figure~\ref{fig:ucrClassificationResults} visualizes the classification performance for all 128 datasets.

\subsection{Inverted Time Warping in $\sphereManifold{1}$}
\label{sec:invertedTimeWarping}


\addtocounter{footnote}{1}
\newcounter{footnoteModifiedGeodesicDistance}
\setcounter{footnoteModifiedGeodesicDistance}{\value{footnote}}
\footnotetext[\value{footnoteModifiedGeodesicDistance}]{%
	Modified version
	utilizing~\eqref{eq:distanceMetricRiemannian}
	for data on Riemannian manifolds.
}

\addtocounter{footnote}{1}
\newcounter{footnoteTTW}
\setcounter{footnoteTTW}{\value{footnote}}
\footnotetext[\value{footnoteTTW}]{%
	Modified version
	utilizing~\eqref{eq:sincInterpolationManifoldTangent}-\eqref{eq:distanceMetricRiemannian}
	for data on Riemannian manifolds.
}

\addtocounter{footnote}{1}
\newcounter{footnoteCPU}
\setcounter{footnoteCPU}{\value{footnote}}
\footnotetext[\value{footnoteCPU}]{%
	CPU used for the experiments is an AMD® Ryzen 5 3500U.
}

\addtocounter{footnote}{1}
\newcounter{footnoteGPU}
\setcounter{footnoteGPU}{\value{footnote}}
\footnotetext[\value{footnoteGPU}]{%
	GPU used for the experiments is an NVIDIA GeForce RTX 4080 Super.
}

Signal alignment is typically performed in an unsupervised setting, 
making it difficult to evaluate the result w.r.t. the optimal solution, 
as it is usually not available for datasets with $N \gg 2$ signals~\cite{Kawano_NTW_2020}.
Here, we mitigate this issue through an inverted time warping approach\footnote{%
	Warping functions that satisfy constraints~\eqref{eq:constraintBoundary}-\eqref{eq:constraintMonton}
	can also be used for creating an artificial temporal shift of a given signal.
	Then, optimal warping functions exist to reverse the tempral shift.
},
which we apply to data in $\sphereManifold{1}$.
Exponential and logarithmic maps are defined as
\begin{gather}
	\label{eq:expmapS3}
	\expmapSD{\dataTangent{}}{\dataManifoldTmp{}}
	=
	\dataManifoldTmp{} \cos(\norm{\dataTangent{}})
	+
	\frac{
		\dataTangent{}
	}{
		\norm{\dataTangent{}}
	}
	\sin(\norm{\dataTangent{}})
	\, ,
	\\
	\label{eq:logmapS3}
	\logmapSD{\dataManifold{}}{\dataManifoldTmp{}}
	=
	\arccos(\dataManifoldTmp{}^{T} \dataManifold{})
	\frac{
		\dataManifold{} - \dataManifoldTmp{}^{T} \dataManifold{} \dataManifoldTmp{}
	}{
		\norm{ \dataManifold{} - \dataManifoldTmp{}^{T} \dataManifold{} \dataManifoldTmp{} }
	}
	\, .
\end{gather}
For the unit sphere $\sphereManifold{D}$,
the tangent space is Euclidean at any base point,
hence the geodesic distance simplifies to
\begin{equation}
	\label{eq:distanceSD}
	\text{dist}_{\sphereManifold{D}}(\dataManifold{}, \dataManifoldTmp{})
	=
	\norm{ \logmapSD{\dataManifold{}}{\dataManifoldTmp{}} }
	=
	\arccos(\dataManifoldTmp{}^{T} \dataManifold{})
	\, .
\end{equation}

More precisely,
starting from a single pre-defined signal~$\dataOriginal{\text{original}}$,
we create $\nbSignals$ synthetic trajectories
$
\dataOriginalSet
=
\{
\dataOriginal{n}
\}_{n=1}^{N}
$
of lengths $\nbData{n} = 100$,
i.e. 
$
\dataOriginal{n}
= 
\{
\dataOriginal{n}[t]
\}_{t=1}^{\nbData{n}}
$,
where $\dataOriginal{n}[t] \in \sphereManifold{1}$.
This is achieved
by generating $\nbSignals$ random warping functions~$\warpingFunction{n}$,
obtained by uniformly sampling
from both~\eqref{eq:warpingFunctionNTW} and~\eqref{eq:warpingFunctionTTW}
with random parameters,
as well as from random spline interpolations of various degrees.
Sampled warping functions that do not satisfy constraints~\eqref{eq:constraintBoundary}-\eqref{eq:constraintMonton} are rejected.
Then,
starting from the original signal~$\dataOriginal{\text{original}}$,
the synthetic signals~$\dataOriginalSet$ are obtained through~\eqref{eq:sincInterpolationManifoldTangent}-\eqref{eq:sincInterpolationManifoldWarped},
resulting in a diverse dataset. 
The goal of this evaluation is to reverse this process, 
by temporally re-aligning the generated data~$\dataOriginalSet$,
i.e., we aim for aligned signals with
$
\dataOriginal{\text{original}} \overset{!}{=} \dataWarped{n}
\,,
\forall\, n \in \curlyBracket{1,\ldots,\nbSignals}
$.

We evaluate RTW in two settings,
with $\nbSignals = 4$ and $\nbSignals = 30$ signals.
For transparent results,
we measure the computation time on a high-grade GPU and a standard CPU.
Furthermore, we evaluate three well-established metrics
\begin{subequations}
	\begin{alignat}{2}
		&
		\label{eq:evalRestorationAccuracy}
		\text{Restoration Accuracy:}
		& \quad
		& \textstyle \sum_{n=1}^{\nbSignals} \dtwDistance{\dataWarped{n}}{\dataOriginal{\text{original}}}
		\\
		&
		\label{eq:evalBarycenterLoss}
		\text{Barycenter Loss:}
		& \quad
		& \textstyle \sum_{n=1}^{\nbSignals} \dtwDistance{\dataOriginal{n}}{\dataMean{}}
		\\
		&
		\label{eq:evalAlignmentQuality}
		\text{Alignment Quality:}
		& \quad
		& \textstyle \sum_{n=1}^{\nbSignals} \dtwDistance{\dataWarped{n}}{\dataMean{}}
	\end{alignat}
\end{subequations}
where~\eqref{eq:distanceSD}
is used within each $\dtwDistanceSymbol$ distance.
The \textit{Restoration Accuracy} describes
how effectively random warping functions are reversed
to restore the original signal~$\dataOriginal{\text{original}}$.
The \textit{Barycenter Loss} measures
how accurately the mean~$\dataMean{}$ of warped signals~$\dataWarped{n}$
represents the data~$\dataOriginal{n}$.
The \textit{Alignment Quality} indicates 
closeness of aligned signals~$\dataWarped{n}$
to their mean~$\dataMean{}$.
The average for each metric is computed from 100 experiments.


\begin{table}[!b]
	\centering
	\caption{Inverted Time Warping on $\sphereManifold{1}$ for $\nbSignals = 4$, $\nbData{} = 100$}
	\label{tab:invertedTimeWarpingNResultSmallN}
	\renewcommand{\arraystretch}{1.2}
	\begin{tabular}{c | c c c c }
		Metric &
		MMDDTW\footnotemark[\value{footnoteModifiedGeodesicDistance}] &
		TTW\footnotemark[\value{footnoteTTW}] &
		NTW\footnotemark[\value{footnoteTTW}] &
		RTW \\
		\hline
		Restoration Accuracy    
		& 0.0601          & 0.1712          &  0.0200          & \textbf{0.0197} \\
		Barycenter Loss         
		& 0.0552          & 0.1801          &  0.0229          & \textbf{0.0206} \\
		Alignment Quality       
		& 0.0403          & 0.0878          &  0.0243          & \textbf{0.0215} \\
		CPU\footnotemark[\value{footnoteCPU}] Runtime in [s] 
		& 3834.3273       & \textbf{3.3734} & 10.7157          & 8.4883          \\
		GPU\footnotemark[\value{footnoteGPU}] Runtime in [s] 
		& 1505.9386       & \textbf{1.5650} &  1.6639          & 1.7574          \\
	\end{tabular}
\end{table}

\begin{table}[!b]
	\centering
	\caption{Inverted Time Warping on $\sphereManifold{1}$ for $\nbSignals = 30$, $\nbData{} = 100$}
	\label{tab:invertedTimeWarpingResult}
	\renewcommand{\arraystretch}{1.2}
	\begin{tabular}{c | c c c c }
		Metric & 
		p-DTW\footnotemark[\value{footnoteModifiedGeodesicDistance}] &
		TTW\footnotemark[\value{footnoteTTW}] & 
		NTW\footnotemark[\value{footnoteTTW}] & 
		RTW \\
		\hline
		Restoration Accuracy    
		& 1.0806  & 0.6458           &   0.0254 & \textbf{0.0293} \\
		Barycenter Loss         
		& 2.1259  & 0.7049           &   0.2954 & \textbf{0.0689} \\
		Alignment Quality       
		& 1.2591  & 0.4332           &   0.3043 & \textbf{0.0749} \\
		CPU\footnotemark[\value{footnoteCPU}] Runtime in [s] 
		& 19.5264 & \textbf{18.8754} & 226.9687 & 103.1399        \\
		GPU\footnotemark[\value{footnoteGPU}] Runtime in [s] 
		& 10.6953 & \textbf{9.4830}  &  10.9298 &  10.2900        \\
	\end{tabular}
\end{table}

\begin{figure}[!t]
	\centering
	\includegraphics[trim=10 0 10 15, clip, width=\linewidth]{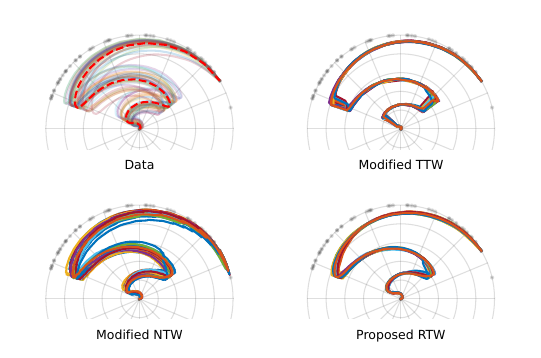}
	\caption{
		Time warping on the $\sphereManifold{1}$ Riemannian manifold.
		All data points lie on the unit circle (grey dots),
		however, for visualization purposes we vary the radius to also indicate the time information between $0$ and $1$.
		Starting from an original signal~$\dataOriginal{\text{original}}$ (dashed line),
		$\nbSignals = 30$ random signals (transparent lines) are generated
		through an inverted time warping approach (see~Sec.~\ref{sec:invertedTimeWarping}).
	}
	\label{fig:invertedTimeWarpingResult}
\end{figure}


\subsubsection{Evaluation for $\nbSignals = 4$}


The literature does not offer time warping schemes for $\nbSignals > 2$ signals
with data on Riemannian manifolds.
Hence, we modify MMDDTW~\cite{Parinya_MMDDTW_2012}
to use the Riemannian distance~\eqref{eq:distanceSD},
as well as TTW and NTW
to adapt~\eqref{eq:sincInterpolationManifoldTangent}-\eqref{eq:distanceMetricRiemannian}.
As Table~\ref{tab:invertedTimeWarpingNResultSmallN} shows, RTW achieves the best performance across all metrics.
While TTW is the fastest, the computational overhead of RTW is well-justified by its results.


\subsubsection{Evaluation for $\nbSignals = 30$}

	
MMDDTW is not capable of handling such a large number of signals in a reasonable time.
Instead, we utilize a pairwise approach by iteratively aligning two signals,
denoted as p-DTW.
The result is shown in
Figure~\ref{fig:invertedTimeWarpingResult} and
Table~\ref{tab:invertedTimeWarpingResult}.
RTW performs best across all metrics
and the difference to NTW becomes more pronounced.


\subsection{Multi-dimensional Riemannian manifolds}


Next,
we extend our experiments to higher-dimensional data
by applying RTW to the Riemannian manifolds~$\sphereManifold{3}$ and~$\spdManifold{D}$.


\subsubsection{Evaluation in $\sphereManifold{3}$}


To further validate our results from section~\ref{sec:invertedTimeWarping},
we compare RTW with QDTW on signals with unit quaternions, i.e., $\sphereManifold{3}$.
Similar to~\cite{Calinon_GaussiansOnRiemannianManifoldsForRobotLearningAndAdaptiveControl_2020},
we artificially generate $\nbSignals = 4$ valid signals of length $\nbData{} = 200$ from the RCFS~\cite{Calinon_RCFS_2023} dataset.
This involves projecting handwritten motion data from $\R^2$ onto the unit sphere of $\sphereManifold{3}$.

In Table~\ref{tab:timeWarpingUnitQuaternionsResults} we observe that the mean signal obtained from RTW
is an improved representation of the original data compared to QDTW.
For the alignment quality, both methods yield similar results,
however, RTW is significantly faster.

\addtocounter{footnote}{1}
\newcounter{footnoteModifiedQDTW}
\setcounter{footnoteModifiedQDTW}{\value{footnote}}
\footnotetext[\value{footnoteModifiedQDTW}]{%
	Modified version
	with $\nbSignals$-dimensional cost matrix as in MMDDTW~\cite{Parinya_MMDDTW_2012}.
}

\begin{table}[!b]
	\centering
	\caption{Time Warping on $\sphereManifold{3}$ for $\nbSignals = 4$, $\nbData{} = 200$}
	\label{tab:timeWarpingUnitQuaternionsResults}
	\renewcommand{\arraystretch}{1.2}
	\begin{tabular}{c | c c }
		Metric            & QDTW\footnotemark[\value{footnoteModifiedQDTW}] & RTW    \\
		\hline
		Barycenter Loss   &         0.0359                          & \textbf{0.0339} \\
		Alignment Quality & \textbf{0.0338}                         & \textbf{0.0338} \\
	\end{tabular}
\end{table}


\subsubsection{Evaluation in $\spdManifold{D}$}


We temporally align signals from the Riemannian manifold of symmetric positive definite (SPD) $D \times D$ matrices.
The exponential and logarithmic maps are
\begin{gather}
	\label{eq:expmapSPD}
	\expmapSPD{\dataTangent{}}{\dataManifoldTmp{}}
	=
	\sqrt{\dataManifoldTmp{}}
	\exp \bracket{
		\sqrt{\dataManifoldTmp{}}^{-1}
		\dataTangent{}
		\sqrt{\dataManifoldTmp{}}^{-1}
	}
	\sqrt{\dataManifoldTmp{}},
	\\
	\label{eq:logmapSPD}
	\logmapSPD{\dataManifold{}}{\dataManifoldTmp{}}
	=
	\sqrt{\dataManifoldTmp{}}
	\log \bracket{
		\sqrt{\dataManifoldTmp{}}^{-1}
		\dataManifold{}
		\sqrt{\dataManifoldTmp{}}^{-1}
	}
	\sqrt{\dataManifoldTmp{}},
\end{gather}
where $\sqrt{\dataManifoldTmp{}}$ and $\sqrt{\dataManifoldTmp{}}^{-1}$ denote
the matrix square root and its inverse, respectively.
Note that the Riemannian metric on the SPD manifold
is Euclidean only at the identity matrix.
Hence, for distance computation under the affine-invariant-Riemannian metric,
we use the canonical form where the Frobenius norm $\norm{\cdot}_F$ yields the geodesic distance
\begin{equation}
	\label{eq:distanceSPD}
	\text{dist}_{\spdManifold{D}}(\dataManifold{}, \dataManifoldTmp{})
	=
	\norm{ \log \bracket{ \sqrt{\dataManifoldTmp{}}^{-1} \dataManifold{} \sqrt{\dataManifoldTmp{}}^{-1} } }_{F}
	\, .
\end{equation}

We compare RTW with two modified versions of MMDDTW.
The first version, denoted as $\text{MMDDTW}_{\manifold}$,
computes the distance between two SPD matrices
according to the geodesic distance defined in~\eqref{eq:distanceSPD}.
The second version, denoted as $\text{MMDDTW}_{\operatorname{C}}$,
utilizes the Cholesky decomposition~$\chol{\cdot}$ to linearize the manifold,
and computes a distance through
\begin{equation}
	\label{eq:distanceMetricCholesky}
	\text{dist}_{\operatorname{chol}}(\dataWarped{}[z] , \dataMean{}[z])
	=
	\norm{
		\chol{\dataWarped{}[z]} - \chol{\dataMean{}[z]}
	}_F
	\, .
\end{equation}

\begin{figure}[!t]
	\centering
	\includegraphics[trim=0 0 0 0, clip, width=\linewidth]{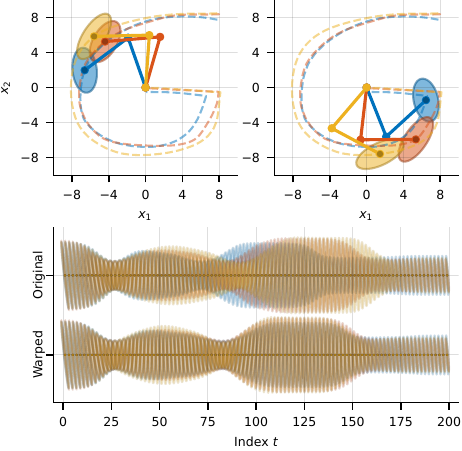}
	\caption{
		A planar robot (solid lines) follows endeffector trajectories (dashed lines), 
		illustrated at $t = 50$ and $t = 100$ for each signal respectively (top-left and top-right).
		The configuration-dependent manipulability is computed for each time step and shown before and after applying RTW (bottom).
	}
	\label{fig:manipulabilityExampleData}
\end{figure}

The data for this experiment was generated from the RCFS~\cite{Calinon_RCFS_2023} dataset,
i.e., a planar robot is controlled to follow the handwritten character G.
We create a dataset for
$\spdManifold{2}$ by computing the manipulability as in~\cite{Jaquier_GaussianMixtureRegressionOnSPDMatrices_2017}, as shown in Figure~\ref{fig:manipulabilityExampleData},
as well as $\spdManifold{8}$ and $\spdManifold{32}$ by computing the mass matrix for a robot with 8 and 32 links, respectively,
resulting in a dataset of $\nbSignals = 3$ signals with a length of $\nbData{} = 200$ each.

We evaluate the Barycenter Loss~\eqref{eq:evalBarycenterLoss} and the Alignment Quality~\eqref{eq:evalAlignmentQuality}
including the geodesic distance given in~\eqref{eq:distanceSPD}.
Table~\ref{tab:timeWarpingManipulabilityResults} presents the results,
supporting our previous findings:
RTW outperforms MMDDTW in terms of Barycenter Loss and alignment quality with remarkable speed.

\begin{table}[!b]
	\centering
	\caption{Time warping with SPD matrices for $\nbSignals = 3$, $\nbData{} = 200$}
	\label{tab:timeWarpingManipulabilityResults}
	\renewcommand{\arraystretch}{1.2}
	\begin{tabular}{ c | c | c c c }
		Metric & D & $\text{MMDDTW}_{\manifold}$ & $\text{MMDDTW}_{\operatorname{C}}$ & RTW     \\
		\hline
		$\text{Barycenter Loss}$   &  2 & 42.1131     & 43.8988      & \textbf{33.5702} \\
		$\text{Alignment Quality}$ &  2 & 42.3271     & 44.6875      & \textbf{23.0621} \\
		\hline
		$\text{Barycenter Loss}$   &  8 & 20.8496     & 21.3290      & \textbf{18.8587} \\
		$\text{Alignment Quality}$ &  8 & 20.9473     & 21.4635      & \textbf{16.6931} \\
		\hline
		$\text{Barycenter Loss}$   & 32 & 22.2595     & 23.3155      & \textbf{20.7979} \\
		$\text{Alignment Quality}$ & 32 & 22.6282     & 23.5338      & \textbf{18.7008} \\
	\end{tabular}
\end{table}


\subsection{Real-Robot Teaching with Time Warping in $\R^{3} \times \sphereManifold{3}$}


\begin{figure*}[tbp]
	\centering
	\includegraphics[
		trim=0 5 0 5, clip,
		width=\linewidth]{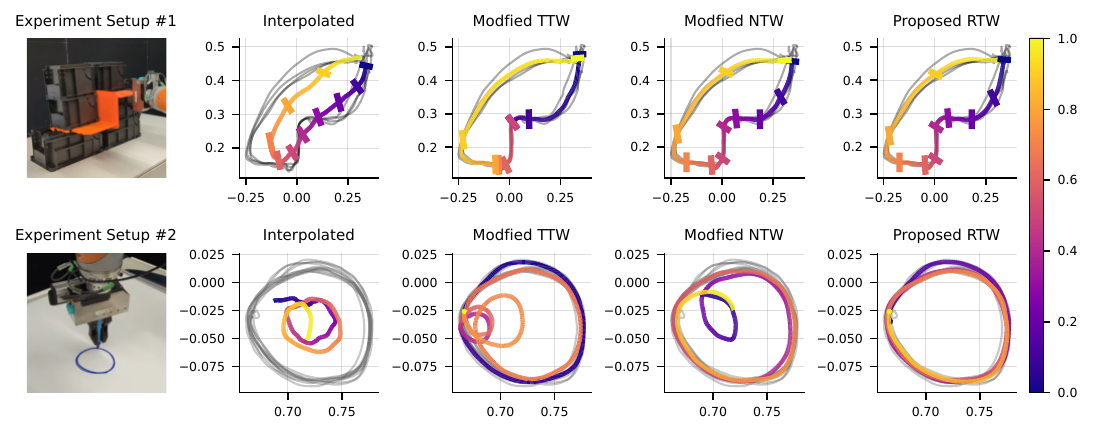}
	\caption{
		Comparison of the results for both robot teaching tasks
		with naive interpolation, modified TTW/NTW and the proposed RTW approach.
		The hand-guided demonstration data~$\dataOriginal{n}$ is drawn in grey and
		the time evolution of the mean signal~$\dataMean{}$ is visualized using the color map.
		Top: the robot has to follow a constrained path, 
		which requires precise positioning and orientation.
		The results are shown from a side perspective and
		the bars on the trajectory indicate the end-effector's orientation (drawn ten times),
		which has to be perpendicular to the motion direction.
		Bottom: the robot has to draw three circles,
		where the difficulty lies in identifying and aligning
		the repetitive characteristics correctly.
		The results are shown from a top-down perspective.
	}
	\label{fig:imitationLearningResults}
\end{figure*}

Finally, we apply time warping in real-robot teaching scenarios,
utilizing a KUKA LBR iiwa manipulator.
We demonstrated $\nbSignals = 4$ signals through hand-guiding control~\cite{Osorio_PhysicalHumanRobotInteractionUnderJointAndCartesianConstraints_2019},
and recorded the end-effector motion at $30$~Hz,
resulting in varying signal lengths $\nbData{n} \in \{2048,\ldots,3667\}$.
Subsequently, the mean signal~$\dataMean{}$ obtained from temporal alignment is executed
employing a state-of-the-art QP-controller~\cite{Bouyarmane_QuadraticProgrammingForMultirobotAndTaskSpaceForceControl_2019},
taking hardware limits and manipulator dynamics into account.

For the trajectory teaching, we denote an end-effector pose as
$
\dataManifold{}
=
\left[ \begin{smallmatrix} 
	\dataManifold{\text{pos}} \\
	\dataManifold{\text{quat}}
\end{smallmatrix} \right]
\in \curlyBracket{\R^{3} \times \sphereManifold{3}}
$,
consisting of position~$\dataPosition \in \R^{3}$
and orientation~$\dataUnitQuaternion \in \sphereManifold{3}$.
Similar to~\cite{Calinon_AnApproachForImitationLearningOnRiemannianManifolds_2017}, 
we concatenate the logarithmic and exponential map into single operations
\begin{gather}
	\label{eq:logmapSE}
	\logmapRdSd{\dataManifold{}}{\dataManifoldTmp{}}
	=
	\left[ \begin{matrix}
		\dataManifold{\text{pos}} \\
		\logmapSThree{\dataManifold{\text{quat}}}{\dataManifoldTmp{}}
	\end{matrix} \right]
	= 
	\left[ \begin{matrix}
		\dataTangent{\text{pos}} \\
		\dataTangent{\text{quat}}
	\end{matrix} \right]
	= \dataTangent{}
	\, ,
	\\
	\label{eq:expmapSE}
	\expmapRdSd{\dataTangent{}}{\dataManifoldTmp{}}
	=
	\left[ \begin{matrix}
		\dataTangent{\text{pos}} \\
		\expmapSThree{\dataTangent{\text{quat}}}{\dataManifoldTmp{}}
	\end{matrix} \right]
	=
	\left[ \begin{matrix}
		\dataManifold{\text{pos}} \\
		\dataManifold{\text{quat}}
	\end{matrix} \right]
	= \dataManifold{}
	\, .
\end{gather}

In this experiment,
we compare RTW
with naive interpolation of the recorded data to the same length,
as well as with modified TTW\footnotemark[\value{footnoteTTW}] and NTW\footnotemark[\value{footnoteTTW}]
utilizing~\eqref{eq:logmapSE} and~\eqref{eq:expmapSE}.
Evaluation is done on two tasks,
(i)~following a constrained path, and
(ii)~repetitively drawing three circles,
both shown in Figure~\ref{fig:imitationLearningResults}.


\subsubsection{Following a constrained path}


We instruct the robot to follow a constrained path with its end-effector,
requiring precise control of position and orientation.
Due to unintended time shifts within the recorded demonstration data, 
naive interpolation does not suffice
and the resulting mean signal 
deviates from the desired path,
resulting in a collision with the environment.
TTW\footnotemark[\value{footnoteTTW}], NTW\footnotemark[\value{footnoteTTW}] and RTW
succeed in following the constrained path without collision,
however,
the warping functions generated from TTW\footnotemark[\value{footnoteTTW}]~\eqref{eq:warpingFunctionTTW}
highly compress the motion in certain sections while stretching others,
resulting in a mean signal that becomes difficult to execute by the robot,
as high velocities are required at the compressed sections.


\subsubsection{Repetitively drawing three circles}


We demonstrate end-effector trajectories consisting of three circles on a whiteboard,
which allows us to investigate the time warping performance
for repetitive segments within provided signals.
Again,
naive interpolation is not sufficient.
Even though both,
TTW\footnotemark[\value{footnoteTTW}] and NTW\footnotemark[\value{footnoteTTW}],
perform better,
they do not align all parts of the trajectory properly.
By only considering the distance between single points,
TTW\footnotemark[\value{footnoteTTW}] and NTW\footnotemark[\value{footnoteTTW}] fail to match repetitive patterns correctly.
This results in severe deviations from the desired motion
at the start and end of the trajectory for both TTW\footnotemark[\value{footnoteTTW}] and NTW\footnotemark[\value{footnoteTTW}].
Furthermore,
TTW\footnotemark[\value{footnoteTTW}] again suffers from fast velocity changes.
Only RTW successfully aligns all parts of the demonstration data.
Repetitive characteristics of the signal are correctly identified due to the Gaussian-weighted window loss~\eqref{eq:gaussianWeightedWindowLoss}
that considers entire segments.
When executing the mean signal~$\dataMean{}$ obtained from the proposed RTW approach,
the robot's end-effector successfully draws the three consecutive circles on the whiteboard,
as shown in Figure~\ref{fig:imitationLearningResults}.

\begin{table}[!b]
	\centering
	\caption{Avgerage Reproduction Error (Constrained Path Task)}
	\label{tab:timeWarpingImitationLearningResultsPath}
	\renewcommand{\arraystretch}{1.2}
	\begin{tabular}{c | c c | c c }
				& \multicolumn{2}{c|}{DTW$_{\text{pos}}$ distance}  &  \multicolumn{2}{c}{DTW$_{\text{ori}}$ distance}        \\
		Method  & Raw Data      & RTW Data           & Raw Data      & RTW Data        \\
		\hline
		DMP     &   0.0392      & \textbf{0.0227}    &   0.0825      & \textbf{0.0371} \\
		GMM     &   0.0302      & \textbf{0.0172}    &   0.0639      & \textbf{0.0334} \\
		GP      &   0.0316      & \textbf{0.0175}    &   0.0695      & \textbf{0.0338} \\
	\end{tabular}
\end{table}	

\begin{table}[!b]
	\centering
	\caption{Avgerage Reproduction Error (Repetitive Circles Task)}
	\label{tab:timeWarpingImitationLearningResultsCircle}
	\renewcommand{\arraystretch}{1.2}
	\begin{tabular}{c | c c | c c }
				& \multicolumn{2}{c|}{DTW$_{\text{pos}}$ distance}  &  \multicolumn{2}{c}{DTW$_{\text{ori}}$ distance}        \\
		Method  & Raw Data      & RTW Data           & Raw Data      & RTW Data        \\
		\hline
		DMP     &   0.0379      & \textbf{0.0222}    &   0.0293      & \textbf{0.0244} \\
		GMM     &   0.0323      & \textbf{0.0220}    &   0.0245      & \textbf{0.0239} \\
		GP      &   0.0329      & \textbf{0.0240}    &   0.0245      & \textbf{0.0235} \\
	\end{tabular}
\end{table}

To further validate the benefits of RTW,
we evaluate its impact on the learning performance across three imitation learning approaches:
Dynamic Movement Primitives (DMP)~\cite{Schaal_DynamicMovementPrimitivesFrameworkForMotorControlInHumansAndHumanoidRobots_2003},
Gaussian Mixture Models (GMM)~\cite{Calinon_GMM_2007},
Gaussian Processes (GP).
Using a leave-one-out cross-validation approach,
we train each method on three demonstrations and evaluate the reproduction quality on the fourth unseen demonstration.
The procedure is repeated four times per task to ensure each demonstration is reproduced once.
Table~\ref{tab:timeWarpingImitationLearningResultsPath} and Table~\ref{tab:timeWarpingImitationLearningResultsCircle}
present the average reproduction errors
for both position and orientation.
The results demonstrate that RTW significantly improves the learning performance across all methods.
This validates that the benefits from RTW extend beyond the time warping process itself
to the entire learning pipeline,
making it a valuable preprocessing tool for imitation learning in robotics.


\section{Conclusion and Future Work}


This letter introduces \emph{Riemannian Time Warping~(RTW)},
a novel approach
that efficiently aligns multiple signals
while considering the geometric structure of the Riemannian manifold the data resides in,
a feature that is essential for many robotic applications.
We conducted extensive experiments on synthetic and real world data, 
proving that RTW consistently outperforms state-of-the-art baselines on both classification and averaging tasks.
Furthermore, we have shown that pre-processing motion data with RTW for training a robot primitive significantly improves the learning result.
The computational complexity of RTW scales linearly with the number of signals and data points, 
which constitutes a significant advancement over Dynamic Programming techniques.

While RTW demonstrates strong performance across various tasks,
there are limitations to be addressed in future work.
RTW has only been evaluated with closed-form exponential and logarithmic maps,
which do not exist for all Riemannian manifolds,
wherefore numerical alternatives may be utilized that result in instabilities.
Additionally, while RTW shows clear benefits for classical motion primitive learning,
it remains unclear whether RTW provides similar advantages for recent imitation learning schemes
like \cite{Lee_MMP_2024} or \cite{Finn_ActAloha_2023},
which encode multi-modal motion strategies based on demonstration data with large spatial variances.


\section*{Acknowledgments}


This work was partly supported by KUKA Deutschland GmbH
and
the state of Bavaria through the OPERA project DIK-2107-0004/DIK0374/01.


\bibliographystyle{ieeetr}
\bibliography{./biblio} 

	
	\vfill
	
\end{document}